# Empirical Evaluation of PDF Parsing and Chunking for Financial Question Answering with RAG

Omar El Bachyr
University of Luxembourg
Luxembourg, Luxembourg
omar.elbachyr@uni.lu

Yewei Song
University of Luxembourg
Luxembourg, Luxembourg
yewei.song@uni.lu

Saad Ezzini
King Fahd University of Petroleum
and Minerals
Dhahran, Saudi Arabia
saad.ezzini@kfupm.edu.sa

Jacques Klein
University of Luxembourg
Luxembourg, Luxembourg
jacques.klein@uni.lu

Tegawendé F. Bissyandé
University of Luxembourg
Luxembourg, Luxembourg
tegawende.bissyande@uni.lu

Anas Zilali
BGL BNP Paribas
Luxembourg, Luxembourg
anas.zilali@bgl.lu

Ulrick Ble
BGL BNP Paribas
Luxembourg, Luxembourg
ulrick.ble@bgl.lu

Anne Goujon
BGL BNP Paribas
Luxembourg, Luxembourg
anne.goujon@bgl.lu

## Abstract

PDF files are primarily intended for human reading rather than automated processing. In addition, the heterogeneous content of PDFs, such as text, tables, and images, poses significant challenges for parsing and information extraction. To address these difficulties, both practitioners and researchers are increasingly developing new methods, including the promising Retrieval-Augmented Generation (RAG) systems to automated PDF processing. However, there is no comprehensive study investigating how different components and design choices affect the performance of a RAG system for understanding PDFs. In this paper, we propose such a study (1) by focusing on Question Answering, a specific language understanding task, and (2) by leveraging two benchmarks from the financial domain, including TableQuest, our newly generated, publicly available benchmark. We systematically examine multiple PDF parsers and chunking strategies (with varied overlap), along with their potential synergies in preserving document structure and ensuring answer correctness. Overall, our results offer practical guidelines for building robust RAG pipelines for PDF understanding.

**ACM Reference Format:**
Omar El Bachyr, Yewei Song, Saad Ezzini, Jacques Klein, Tegawendé F. Bissyandé, Anas Zilali, Ulrick Ble, and Anne Goujon. 2026. Empirical Evaluation of PDF Parsing and Chunking for Financial Question Answering with RAG. In *2026 IEEE/ACM 48th International Conference on Software Engineering (ICSE-SEIP '26), April 12–18, 2026, Rio de Janeiro, Brazil*. ACM, New York, NY, USA, 12 pages. https://doi.org/10.1145/3786583.3786911

## 1 Introduction

PDFs provide a standard way to preserve the layout of documents, making them ideal for official disclosures. Unlike structured formats, PDF files encapsulate data in a format meant for human consumption. However, the heterogeneous mix of content in PDFs, including text, tables, images, and forms, leads to difficulties in parsing and leveraging this information effectively for automated systems. Several studies [5, 30] have highlighted the inherent difficulties of extracting information from PDFs in these contexts, including challenges with inconsistent formatting, complex table structures, and the need for advanced parsing techniques. These challenges make PDF understanding a critical bottleneck in all document-related tasks, including Question Answering (QA), document understanding (DU), information retrieval (IR), and summarization, especially in industry settings where automated document workflows must be both accurate and scalable.

Retrieval-Augmented Generation (RAG) technique [26] has emerged as a promising solution to tackle such tasks. A typical RAG comprises two stages: an offline pre-processing phase and an online query phase. In the offline phase, source documents are parsed and then chunked (i.e., split into logical text segments such as paragraphs), embedded, and then ingested into a vector store. In the online phase, when a user asks a question, the RAG system first searches a collection of chunks relevant to the asked question and then provides those documents to an LLM to generate a precise final answer. Chunking documents is necessary because both embedding models and language models have a limited context window, that is, a maximum number of tokens they can process at once, making it impossible to embed and submit an entire document as input. This two-stage approach significantly improves factual accuracy and reduces hallucinations, as the LLM's output is anchored in the retrieved content. However, the effectiveness of a RAG pipeline depends critically on how the PDFs are parsed and chunked in the first place. If parsing produces incorrect structure (e.g., misaligned tables or merged columns) or chunking disrupts logical units (e.g., splitting tables or detaching figures from captions), the retrieval component may fail to retrieve the information for correct answers.

Although RAG performance directly depends on parsing and chunking during pre-processing, there is a lack of prior work that jointly evaluates PDF parsing and chunking strategies for specific tasks such as QA. Most existing research [1, 45] has tackled pieces of the problem in isolation. For example, in [1], the authors conducted a comprehensive tool-agnostic evaluation of ten popular PDF-parsing frameworks across six diverse document categories, such as financial reports, scientific papers, patents, or law & regulation documents. In a different line of work, [45] introduced enhanced chunking strategies, segmenting documents by their structural components (headings, tables, etc.) instead of relying on uniform paragraph divisions.

However, to the best of our knowledge, no prior study has systematically combined and compared different document parsers and different chunking approaches within a unified framework. This leaves an important gap: practitioners lack guidance on how the choice of PDF parsing tool, which affects text and table extraction, and how documents are chunked, which boosts retrieval but can break context or increase processing time, together influence end-to-end answer quality. A holistic evaluation is needed to understand these interactions, especially for complex documents that demand, for instance, both textual and tabular comprehension.

To address this gap, this work presents an empirical study assessing the performance of various PDF parsing and chunking strategies in an RAG-based pipeline. For this study, we focus on a specific document category -financial documents- and a specific task —Question Answering (QA)-. This key natural language understanding task perfectly mirrors analysts' real-world workflow, which is locating relevant and precise information and reasoning over it to generate accurate answers, enabling us to evaluate the full RAG pipeline (parsing, chunking, retrieval, and generation). We conduct rigorous experiments on two complementary benchmarks: FinanceBench [19] for text-based QA and our newly generated TableQuest for table-focused QA. FinanceBench is a recently introduced test dataset for open-book question answering about publicly traded companies, featuring 150 text-based questions with associated evidence from annual reports. In contrast, TableQuest (see Section 3) is a test set of 116 QA pairs that evaluate the RAG system's ability to retrieve and reason over tabular data in the context of financial reports.

*What makes financial documents particularly suitable targets for this study?* In the financial domain, a significant portion of valuable information and data is encoded in PDF files. Financial reports, regulatory filings, earnings statements, contracts, and other critical documents are often distributed in PDF format, making it one of the most prevalent data sources in the industry. Ensuring an accurate interpretation of these documents is critical, as even minor errors or misreadings can lead to substantial financial losses or regulatory non-compliance [15, 17]. This high-stakes context motivates robust PDF understanding methods for financial applications, where analysts and automated systems alike must reliably extract facts and figures from complex reports. Our contributions are as follows:

(1) **Evaluation of PDF parsers and chunking strategies.** We benchmark multiple PDF parsers on datasets of real-world financial documents, recording processing metrics such as parsing time, and for each parser, evaluating various chunking strategies by reporting the number of generated chunks.

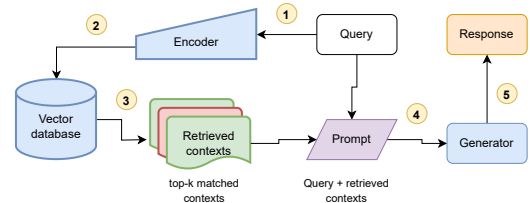

Figure 1: An overview of the Retrieval-Augmented Generation (RAG) architecture.

(2) **TableQuest dataset.** We build TableQuest, our automatic-verified QA benchmark derived from the tables FinanceBench PDFs. We detail its automatic generation, annotation protocol, and dataset statistics (Section 3).

(3) **Comprehensive RAG benchmarks.** We evaluate end-to-end RAG pipelines, varying parsers, chunking strategies, and LLMs, on both FinanceBench and TableQuest, demonstrating the impact of each component choice on retrieval accuracy and answer correctness (Section 5).

All code, prompts, datasets, and experiment configurations are publicly available on **our GitHub repository**, supporting reproducibility and transfer to practice.

## 2 Background

**Retrieval-Augmented Generation (RAG)** is a paradigm that enhances a generative language model by incorporating an external retrieval mechanism, allowing models to produce accurate, evidence-grounded responses by dynamically consulting external documents [26]. A typical RAG pipeline, as illustrated in Figure 1, involves three main components: an embedding model, a vector database, and a response generator.

In an *offline indexing phase*, all source documents are converted into embeddings by the embedding model and ingested into the vector database. At *inference time*, the user's query is embedded using the same embedding model in the offline phase (1→2) and used to retrieve the top-$k$ most similar chunks from the database (3); these retrieved contexts are then fed alongside the user query into an LLM as a response generator (4) to produce the final answer (5). In other words, RAG combines a parametric memory (the knowledge stored in the model weights) with a non-parametric memory (external documents or data) to produce answers. RAG addresses limitations of standard language models, such as hallucinations and outdated knowledge, by providing up-to-date, verifiable context, thus enhancing both factual accuracy and credibility [13]. Initially applied to open domain question answering and dialogue systems [8, 26], RAG has become particularly valuable for document question answering, where efficient *parsing* and *chunking strategies* are crucial to retrieve relevant passages from a large document corpus.

**PDF Parsing.** PDF stands for Portable Document Format, a file format created by Adobe Systems in 1992 and later standardized as ISO 32000 [2]. It encapsulates a fixed-layout flat document: text, fonts, vector graphics, raster images, and other information so that documents appear the same regardless of software, hardware, or operating system. PDF parsing involves extracting this structured content from PDF. Because PDFs store text as positioned glyphs without inherent semantic markup, parsers must infer logical structure (paragraphs, tables, headings) using geometric heuristics or machine learning, which becomes especially challenging for complex layouts, multi-column text, and scanned images requiring OCR. A range of tools addresses these needs, from fast, baseline libraries like PyPDF2 and PyMuPDF to more sophisticated frameworks such as pdfminer, pdfplumber, and Unstructured that offer fine-grained layout analysis and table detection, each balancing trade-offs between speed, accuracy, and implementation complexity [1].

**Text Chunking.** Even as cutting-edge LLMs are being built with very long context windows [38], studies find that retrieval augmentation remains advantageous for both accuracy and efficiency [43]. Moreover, the LLM ability to process longer inputs significantly increases computational demands due to the quadratic complexity of the self-attention mechanism [23]. This creates challenges for computational efficiency, particularly in terms of speed and memory, which are especially critical for Retrieval-Augmented Generation (RAG) tasks. To address these challenges, it is essential to provide LLMs with only the most relevant and accurate information during the retrieval phase. Effective information retrieval relies on two key factors: accuracy, which involves selecting the most relevant data, and granularity, which ensures that the extracted content is appropriately sized. For example, extracting overly large chunks of irrelevant data introduces noise that can impair the model's performance. Additionally, embedding models compress large volumes of information into dense vectors, but this process is inherently lossy. Thus, vector representations of text are most effective when they encode singular, well-defined concepts rather than multiple, unrelated ideas. To reduce noise and maintain high-quality embeddings, chunking is crucial. By dividing the text into smaller, coherent units, chunking ensures that the retrieved data is precise and relevant. This leads to better retrieval performance and higher-quality outputs in RAG systems, making chunking a fundamental component of modern retrieval methods.

Several studies have emphasized the importance of chunking strategies in enhancing RAG performance [45], as the choice of chunk size and overlap can significantly influence context retention and retrieval quality. Poor chunking choices can lead to loss of context, incomplete information, and reduced performance in downstream tasks such as Financial PDF Question Answering.

Although chunking is often the first step in data ingestion for RAG pipelines, there is limited literature on evaluating different chunking strategies.

**FinanceBench Dataset.** FinanceBench [19] is a benchmark for evaluating LLMs on financial QA over PDFs. It comprises 10,231 question–answer pairs derived from 361 public filings of four document types: annual reports (Form 10-K), quarterly reports (Form 10-Q), current event reports (Form 8-K), and earnings reports. These reports cover 40 publicly traded U.S. companies from 2015 to 2023. Each entry includes a question, its answer, supporting evidence, and metadata such as company name, sector, and document details. However, only a subset of 150 QA pairs over 84 PDFs is publicly available in the original paper. The dataset is intended to reflect actual analyst research needs and organizes its questions into three categories:

- **Domain-relevant questions:** Generic finance queries vetted by analysts (e.g., dividend history, margin consistency).
- **Novel generated questions:** Company and report-specific prompts crafted by experts to require multi-step reasoning and mirror real-world scenarios.
- **Metrics-generated questions:** Template-driven items that compute or extract base and derivative financial metrics from multiple statements, blending simple lookups with basic calculations.

Table 1 presents key statistics for each FinanceBench document type, while Table 2 presents statistics on the QAs.

**Table 1: FinanceBench PDF statistics by document type.**

| Document Type | Total Docs | Total Pages | Mean Pages |
|---|---|---|---|
| 10k | 64 | 10,757 | 168 |
| 10q | 8 | 906 | 113 |
| 8k | 6 | 248 | 41 |
| Earnings | 6 | 102 | 17 |
| **All Documents** | **84** | **12,013** | **143** |

**Table 2: FinanceBench QA statistics (Avg. lengths in tokens).**

| Category | QA Pair Count | Avg. Answer Length | Avg. Question Length |
|---|---|---|---|
| domain-relevant | 50 | 26.98 | 33.46 |
| metrics-generated | 50 | 56.66 | 3.96 |
| novel-generated | 50 | 22.40 | 21.82 |
| **Total** | **150** | – | – |

## 3 TableQuest: Challenging Table-Based QA for Retrieval-Augmented Generation

FinanceBench evaluates a model almost exclusively on text-based questions that can be answered by scanning narrative text blocks within a PDF page, so a system can excel without accurately processing the tables that contain the most important financial data. Ignoring tables leaves out two key skills: (1) accurately locating the page that holds the relevant table in a long report, and (2) reasoning over tabular layouts whose rows, columns, and number relationships differ markedly from plain text. Focusing on tables is vital because it checks if a model can read complex layouts, perform calculations on table cells, and draw insights that text alone cannot give.

To close this gap, we propose TableQuest, a benchmark that shifts the evaluation target from narrative paragraphs to table-focused pages, crafting each query so that success depends on retrieving the correct table and reasoning over its structure and values. The benchmark presents questions of increasing difficulty, starting with simple single-table queries and progressing to cross-table synthesis. In summary, TableQuest complements FinanceBench by testing the very skills that text-only benchmarks overlook, providing a more comprehensive assessment of real-world financial QA competence. TableQuest includes three types of carefully constructed questions so that, by answering them correctly, the models demonstrate the following capabilities:

- **Easy – Single-table Extractive:** Retrieve the page, locate a small table (5–10 cells), and reproduce an exactly stated value, demonstrating basic cell-level lookup and comprehension.

- **Medium – Single-table Numerical:** Retrieve the page, locate a medium-sized table (50–100 cells), then compute a value, such as a sum, ratio or growth rate, from numbers in one table.
- **Hard – Multi-table Analytical:** Retrieve a page containing multiple large tables and synthesise figures across them to produce an insight.

Figure 2 summarises the four-stage construction pipeline that we followed to build TableQuest. Starting from the same 84 FinanceBench PDFs, we (i) detect and count tables on every page, (ii) filter pages with at least one table and classify them by table count and cell count, (iii) perform stratified sampling to obtain 70 pages per difficulty tier while preserving industry and year diversity, and (iv) generate question–answer pairs via GPT-4o with chain-of-thought prompting, followed by meticulous automatic verification. We now detail each of these four steps.

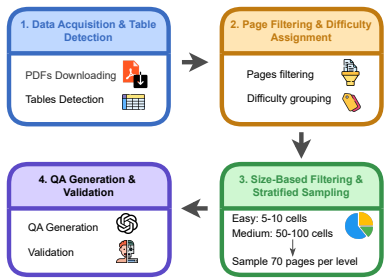

**Figure 2: TableQuest Dataset Construction Process.**

**1. Data Preparation & Table Detection.** The data preparation stage begins with finding and downloading the 84 FinanceBench PDFs. We then apply a layout detection model [37] on every page, filtering out those without any existing tables, and retaining only pages that contain at least one table.

**2. Page Filtering and Difficulty Assignment.** Next, we automatically categorize the remaining pages into two broad difficulty groups based on the number of detected tables: (1) Easy/Medium pages group, each containing exactly one table, and (2) Hard pages group, containing two or more tables.

**3. Size-Based Filtering & Stratified Sampling.** To ensure diversity and reduce bias, we perform stratified sampling within each difficulty level, selecting 70 pages per level. Within the Easy/Medium cohort, we further categorize pages by table size: those with 5–10 filled cells form the Easy subset, emphasizing simple lookups, while those with 50–100 filled cells comprise the Medium subset, focusing on moderate numerical computations. Finally, a manual review phase removes any pages that do not strictly adhere to our criteria.

**4. QA Generation and Validation.** This stage consists of two main steps. First, we use pages from the previous step to generate QA pairs for each difficulty level using GPT-4o [18], a proprietary vision-language model with strong visual understanding. For each difficulty level (easy, medium, hard), we crafted a tailored chain-of-thought [42] (CoT) prompt (see Github repository) to encourage the model to reveal its reasoning process and improve answer quality. Second, we apply automatic validation using three proprietary LLMs with strong multimodal (text & image) performance: GPT-5, o3, and GPT-4.1. Each model independently judges the QA pair given the page as input, and a majority vote decides whether it is retained or rejected, ensuring consistency and high dataset fidelity. The resulting dataset statistics, including counts, average lengths, and retention rates, are summarized in Table 3.

Through a rigorous, multi-stage workflow, comprising automated table detection, page filtering, difficulty grouping, stratified sampling, QA generation with chain-of-thought and automatic validation, we curated TableQuest, a robust dataset of table-centric financial QA pairs for evaluating retrieval-augmented generation systems.

**Table 3: TableQuest QA statistics after automatic validation (Avg. lengths in tokens).**

| Category | QA Pairs | Avg. Q Len | Avg. A Len | Retained | Unanimous (3C–0I) |
|---|---|---|---|---|---|
| Easy | 56 | 18.10 | 3.24 | 96.6% | 94.8% |
| Medium | 31 | 24.72 | 2.81 | 58.5% | 49.1% |
| Hard | 29 | 50.29 | 54.60 | 50.0% | 37.9% |
| **Total** | **116** | – | – | – | – |

## 4 Empirical Study Methodology

In this section, we detail the methodology we follow to conduct our empirical evaluation, which aims at comparing the performance of various PDF parsing and chunking strategies in an RAG-based pipeline. We will first announce the research questions we answered in this study (Section 4.1). We then present the benchmarking pipeline we followed to conduct the study (Section 4.2), the PDF parsing (Section 4.3) and text chunking (Section 4.4) used, and the experimented embedding models (Section 4.5). We finally detail the experimental design and evaluation of the study (Section 4.6).

### 4.1 Research Questions

Our empirical study focuses on how PDF parsing and chunking choices shape both retrieval and answer-generation performance in a RAG pipeline over financial PDFs. To comprehensively characterize each stage, we ask:

**RQ1:** Which retriever family (keyword, dense, late-interaction, or sparse-hybrid) yields the strongest page-level retrieval for narrative and table queries?

**RQ2:** How does the choice of PDF parser affect page-level retrieval performance?

**RQ3:** How do chunking strategy and overlap percentage influence retrieval accuracy at the page level?

**RQ4:** Do specific combinations of PDF parsers and chunking strategies synergistically enhance performance?

**RQ5:** How does LLM size impact the quality and correctness of generated answers in this RAG setting?

### 4.2 Overall Benchmarking Pipeline

Figure 3 illustrates our end-to-end PDF-QA benchmarking pipeline, repeated for every parser–chunker–indexer variant. This pipeline is composed of three steps: Offline, Online and Evaluation. In the **offline** stage, financial report pages from FinanceBench and our newly introduced dataset TableQuest are processed by one of the parsers (Section 4.3), then split into chunks under one of the chunking strategies described in Section 4.4, and finally indexed by each embedding model (see Section 4.5) to build separate vector stores. In the **online** stage, each dataset's query (from FinanceBench or

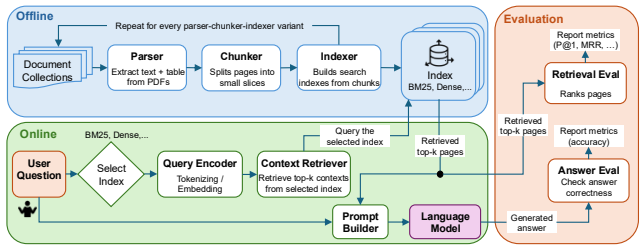

Figure 3: End-to-end PDF-QA pipeline.

Tablequest) is encoded and queried against the selected index to retrieve the top-$K$ relevant *pages*. Those pages, together with the original question, form the prompt for one of several tested LLMs (see Table12), which generates an answer. In the **evaluation** stage, we (a) assess retrieval quality by measuring the pipeline's ability to return the correct pages for each query, and (b) assess the accuracy of the answers by comparing the generated answers with the ground truth answers in both datasets. This modular design cleanly isolates the effects of parsing, chunking, indexing, and model choice on both retrieval and generation performance.

### 4.3 PDF Parsing

For our experiments, we use the six widely adopted open source PDF parsers that are listed in Table 4. These parsers represents the primary approaches used in practice, from lightweight, pure-Python libraries (PyPDF2, PyMuPDF, pdfminer.six, pypdfium2) to more layout-aware tools (pdfplumber) and OCR-driven, model-based systems (Unstructured). Together they span rule-based extraction, table and structure handling, and full document analysis, giving us a broad view of how different parsing capabilities affect downstream chunking and retrieval. By evaluating each one under the same conditions, we ensure our findings apply across the spectrum of commonly used PDF processing tools.

Each parser is applied to all 337 PDF pages (168 from FinanceBench and 169 from TableQuest), each page linked to at least one QA pair. Here are brief summaries of the six parsers:

Table 4: Parsers Comparison on PDF Data Extraction.

| Parser | Version | Extractable Data Types | Output Type | Extraction Method |
|---|---|---|---|---|
| PyPDF2 | 3.0.1 | T - M - I | Text | Rule-Based |
| PyMuPDF | 1.25.5 | T - M - I - TAB | Text | Rule-Based |
| pdfminer.six | 20250327 | T - M - I | Text | Rule-Based |
| pdfplumber | 0.11.6 | T - M - I - TAB | Text, CSV | Rule-Based |
| pypdfium2 | 4.30.1 | T - M - I | Text | Rule-Based |
| unstructured | 0.17.2 | T - M - I - TAB - DS | Text, JSON, Markdown, Markdown, CSV, HTML | Rule-Based, OCR, Model-Based |

**Abbreviations.** T = Text; M = Metadata; I = Images; TAB = Tables; DS = Doc Struct.

**PyPDF2 [33]** A pure-Python library that enables operations such as splitting, merging, cropping, and transforming PDF pages. It also supports text and metadata extraction, as well as adding custom data, viewing options, and passwords to PDF files.

**PyMuPDF [4]** is a high-performance Python binding to the MuPDF C/C++ engine, designed for fast data extraction, conversion, and manipulation of PDF and other documents (e.g., XPS and EPUB). Provides functionalities for text and image extraction, document rendering, and manipulation of annotations and form fields.

**pdfminer.six [32]** is a community-maintained fork of the original pdfminer library, focused on precise, rule-based extraction of text, fonts, and layout data directly from PDF source streams. It provides access to character coordinates, font information, and color data, and can output text to formats like HTML or hOCR.

**pdfplumber [21]** builds on pdfminer.six to offer a user-friendly API for *plumbing* PDFs: retrieving every character, word, line, and rectangle, and facilitating visual debugging of extraction processes. It adds table-extraction methods and preserves column layouts via customizable parameters, is optimized for machine-generated PDFs.

**pypdfium2 [34]** is a Python wrapper for Google's PDFium engine, offering fast page rendering plus text and image extraction, with active maintenance.

**Unstructured [40]** is an open source Python library aimed at simplifying the ingestion and pre-processing of unstructured data from various formats such as PDFs, HTML, Word documents, and images. Provides modular components for transforming unstructured data into structured outputs, enhancing workflows for LLMs.

Since financial PDFs often embed critical information in tables, it is important to note that these parsers vary in how they handle tables. It's notable that only PyMuPDF, pdfplumber, and Unstructured provide built-in support for table extraction, while the others handle just text (including table), metadata, and images (third column in Table 4). This distinction lets us evaluate whether native table extraction yields better downstream chunking and retrieval performance.

### 4.4 Text Chunking

To make each parsed page retrievable while still giving the LLM enough context, we segment text with six chunking strategies [1] that are among the most widely used in retrieval-augmented generation pipelines: *token*, *sentence*, *semantic*, *recursive*, *SDPM*, and *neural*. All strategies respect a hard cap of 512 tokens (i.e., the size of each chunck is 512 tokens maximum), yet differ in how they place boundaries:

**Token chunking** strategy involves dividing text into segments based on a specified number of tokens (512 in our case), ensuring that each chunk stays within defined token limits.

**Sentence chunking** is a content-aware chunking strategy that cuts documents only at sentence boundaries, optionally grouping a fixed number of adjacent sentences so every resulting chunk is a complete, self-contained unit.

**Recursive chunking** strategy involves dividing a document into smaller segments by sequentially splitting it into specific delimiters, such as paragraphs, sentences, and words, until each segment reaches the desired size (512 tokens maximun).

**Semantic chunking** strategy segments the text by evaluating the semantic similarity between sentences using an embedding model. When the difference between embeddings of consecutive sentences exceeds a predefined threshold, a new segment is initiated.

**Semantic Double-Pass Merging (SDPM) chunking** strategy enhances semantic chunking by employing a two-phase merging process. Initially, it groups content based on semantic similarity. Subsequently, it merges these similar groups within a defined skip

---
[1]All chunkers are provided by the open-source Chonkie library: https://github.com/chonkie-inc/chonkie

window, effectively connecting related content that may not be sequential in the text.

**Neural chunking** is a model-driven, content-aware strategy that asks a fine-tuned transformer (usually BERT) to spot where the meaning of a document changes and to cut only at those boundaries, producing a variable length but coherent chunks.

By default, we also apply a 25% overlap (i.e., each new chunk reuses the last 128 tokens of its predecessor), so that key information is not cut in half while keeping the total number of chunks reasonable. When examining pure parser–chunker interactions (RQ4 in Section 5), we instead fix overlap at 0% so that any performance differences reflect only boundary placement.

### 4.5 Experimented Embedding Models

To ensure a thorough evaluation, we selected the encoders listed in Table 5, spanning the full spectrum of embedding strategies. We include Okapi BM25 [35], relying on pure lexical overlap, alongside ColBERT [24], which applies contextualized late-interaction for token-level matching, E5 [41], providing fixed-length dense vectors across dozens of languages, and SPLADE [25], which generates interpretable sparse activation representations.

**Table 5: Summary of Retrieval Model Configurations and Sizes (Millions of Parameters)**

| Model | Type | Params (M) | Max Seq. Len | Embedding Dim. |
|---|---|---|---|---|
| Okapi BM25 [35] | Keyword-based | — | N/A | N/A |
| SPLADE-v3 [25] | Sparse | 284 | 512 | 768 |
| E5-large [41] | Dense (single-vector) | 560 | 512 | 1024 |
| ColBERT [24] | Dense (multi-vector) | 110 | 512 | 128 |

### 4.6 Experimental Design & Evaluation

*4.6.1 Datasets.* We evaluate retrieval on the two financial-domain QA benchmarks introduced earlier. **FinanceBench** offers 150 narrative-text questions, whereas our **TableQuest** corpus (see Section 3) adds 164 table-focused QA pairs drawn from a subset of the same PDF reports. For both datasets we index only those pages that are linked to at least one question.

*4.6.2 Metrics.* We evaluate page–level retrieval with three complementary measures. Precision@1 captures how frequently the very first page returned is relevant, reflecting the user's immediate experience. Recall@3 checks whether all ground-truth pages appear anywhere in the top three results, emphasising coverage. Mean Reciprocal Rank (MRR) aggregates precision and rank across the entire result list, rewarding systems that place relevant pages as high as possible.

Answer correctness is evaluated automatically using GPT-4o-mini as an LLM-as-judge. After context retrieval and answer generation, the judge compares the model's answer with the dataset's reference answer and issues a pass/fail accuracy label according to standard QA grading guidelines.

By pairing objective retrieval scores with semantic answer grading, we can disentangle the contribution of each pipeline component to overall QA performance.

*4.6.3 Hardware & Software.* We implemented our local RAG pipeline using a combination of open-source libraries: rank-bm25 [29] for BM25 indexing, sparsembed [36] for SPLADE, PyLate [20] for ColBERT-style indexing, and HuggingFace Transformers [39] for dense retrieval. Prompt assembly and answer generation are handled locally via Ollama [31], and retrieval performance is measured with the ir-measures [22] library. All experiments were conducted on a workstation equipped with an NVIDIA RTX 5000 Ada GPU (32 GB VRAM).

## 5 Experimental Results

With the setup described earlier, we can now explore answers to various research questions concerning our RAG pipeline over financial PDFs.

> **RQ1:** Which retriever family (keyword, dense, late-interaction, or sparse-hybrid) provides the strongest page-level retrieval for text and table QA in financial PDFs?

**Test:** *We evaluate the four embedding models from HuggingFace listed in Table 5 on FinanceBench (text QA) and our new table-focused QA dataset TableQuest. Retrieval performance is measured by P@1, R@3, and MRR. Text is extracted using six PDF parsers (pdfminer, PyMuPDF, PyPDF2, Unstructured, pdfplumber, pypdfium2) and split with six chunking strategies (token, sentence, semantic, recursive, SDPM, neural). To ensure fair comparisons, we fix the chunk size at 512 tokens with 25% (128 tokens) overlap to provide sufficient context while minimizing fragmentation effects resulting from text chunking.*

**Table 6: BM25, E5, ColBERT, SPLADE retrieval on FinanceBench & TableQuest (avg. P@1, R@3, MRR).**

| Retriever | FinanceBench | | | TableQuest | | |
| | P@1 | R@3 | MRR | P@1 | R@3 | MRR |
|---|---|---|---|---|---|---|
| BM25 [35] | 0.129 | 0.212 | 0.210 | 0.576 | 0.757 | 0.677 |
| E5 [41] | **0.568** | **0.763** | **0.700** | 0.651 | 0.824 | 0.751 |
| ColBERT [24] | 0.479 | 0.652 | 0.610 | **0.760** | **0.919** | **0.844** |
| SPLADE [25] | 0.473 | 0.674 | 0.625 | 0.683 | 0.855 | 0.784 |

As shown in Table 6, BM25 [35] only occasionally retrieves relevant pages for queries, achieving P@1 of 0.129 and 0.576 on FinanceBench and TableQuest, respectively, with correspondingly low R@3 and MRR values. This finding confirms that keyword matching alone is insufficient to capture the deep semantics required by financial narratives or structured table queries.

In the FinanceBench dataset, E5 [41] outperforms all other methods with P@1 of 0.568, R@3 of 0.763, and MRR of 0.7, indicating that a single-vector dense bi-encoder effectively models the global semantics of text-focused financial queries. In contrast, in our TableQuest table-focused benchmark, ColBERTRetriever [24] achieves the highest P@1 (0.76), R@3 (0.919) and MRR (0.844), demonstrating that its token-level late interaction mechanism excels at matching query tokens to specific table elements.

SPLADE [25] occupies the second rank in both benchmarks, providing a robust compromise by combining sparse, learned term expansions with contextual embeddings. Its performance remains consistently strong across mixed query types, suggesting that sparse-hybrid models can effectively handle workloads that include both narrative text and structured tables.

These results illustrate that retriever selection should be guided by query characteristics: dense bi-encoders are recommended for narrative financial texts, late interaction models for structured table retrieval, and sparse-hybrid retrievers when both text and table understanding are required. All metrics reported here refer to page-level retrieval performance under a maximum of three relevant pages per query.

**RQ1 Short Answer.** Dense retrievers excel on narrative text queries, late-interaction models lead on table queries, and sparse-hybrid retrievers deliver the best overall balance.

### RQ2: How do different PDF parsers affect page-level retrieval performance on financial documents?

**Test:** *Using SPLADE as the fixed retriever (the sparse-hybrid model shown in RQ1 to deliver the best overall balance between text-focused and table-focused queries), we examine how six PDF parsers (pdfminer, PyMuPDF, PyPDF2, Unstructured, pdfplumber, pypdfium2) affect page-level retrieval on FinanceBench and TableQuest. Retrieval is evaluated by P@1, R@3, and MRR under a full-factorial design across six chunking strategies (token, sentence, semantic, recursive, SDPM, neural), with 512 tokens chunks and 25% overlap (128 tokens). To isolate parser effects, we average each parser's scores over all chunkers. In two additional sub-research questions, we also explore the stability of the pasers (RQ2b) as well as the runtime performance of the parsers (RQ2c).*

**RQ2a. Which PDF parser yields the best average page-level retrieval?** As Table 7 shows, on FinanceBench, where queries target narrative financial text, pdfminer's careful stream extraction yields the highest page-level retrieval (MRR 0.646), edging out pdfplumber by 1.9% (0.634 → 0.646). It also surpasses unstructured's OCR-based approach (MRR 0.630) by 2.5%, demonstrating that even layout-agnostic extraction can closely match OCR-based extraction on plain text. The mid-tier parsers, pypdfium2 (0.628) and PyMuPDF (0.626), show robust general text handling, while PyPDF2, using simpler heuristics, lags behind with an MRR of (0.588).

As for table-focused queries in TableQuest, pdfplumber moves to the top (MRR 0.807), reflecting its cell grouping and table detection strengths. pdfminer remains competitive at 0.793, suggesting that its stream layout is still useful for structured pages. Unstructured's OCR-based layout (0.788) ranks third, followed closely by PyMuPDF (0.782) and pypdfium2 (0.777), with PyPDF2 again finishing last (0.758), indicating that it struggles more to preserve structured table information.

Overall, retrieval results imply that pdfminer is the best default for narrative financial documents, whereas pdfplumber is preferable when tables dominate. OCR-based parser (unstructured) offers a robust middle ground, and modern render-based extractors (PyMuPDF, pypdfium2) remain competitive, but parsers lacking advanced layout logic (PyPDF2) risk noticeable accuracy loss across both query types.

**RQ2b. How stable is each PDF parser's performance across chunking strategies?** Figure 4 details the variation of the performance score of the six parsers when the chunking strategy varies. On FinanceBench, retrieval stability varies widely (Figure 4a). pdfminer shows the narrowest error bars (std = 0.019), closely matched by Unstructured (std = 0.020), indicating that their MRR hardly shifts when you change chunk boundaries. pypdfium2 (std

**Table 7: Parser impact on page-level retrieval (512-token chunks, 25% overlap). Rankings by avg. MRR across datasets.**

| | FinanceBench | | | TableQuest | | | |
|---|---|---|---|---|---|---|---|
| Parser | P@1 | R@3 | MRR | P@1 | R@3 | MRR | Rank |
| pdfplumber | 0.477 | 0.692 | <u>0.634</u> | 0.710 | 0.879 | **0.807** | 1 |
| pdfminer | 0.501 | 0.696 | **0.646** | 0.701 | 0.859 | <u>0.793</u> | 2 |
| unstructured | 0.472 | 0.688 | 0.630 | 0.681 | 0.876 | 0.788 | 3 |
| pymupdf | 0.474 | 0.668 | 0.626 | 0.682 | 0.842 | 0.782 | 4 |
| pypdfium2 | 0.470 | 0.685 | 0.628 | 0.672 | 0.852 | 0.777 | 5 |
| pypdf2 | 0.442 | 0.614 | 0.588 | 0.651 | 0.822 | 0.758 | 6 |

= 0.036) and PyPDF2 (std = 0.032) exhibit the greatest swings (up to ±0.03 MRR), so their performance depends heavily on how text is chunked. PyMuPDF (std = 0.024) and pdfplumber (std = 0.025) fall in between, with moderate sensitivity to the chunking strategy.

On TableQuest, overall variance is relatively smaller (Figure 4b). pdfminer again is most stable (std = 0.012), followed by pypdfium2 (0.013). pdfplumber and unstructured show slightly larger spreads (0.016), while PyPDF2 and PyMuPDF (0.018 and 0.022) are the most affected by the chunking strategy in table-rich pages.

Taken together, pdfminer and pdfplumber deliver not only high average MRR (RQ2a) but also the greatest robustness to chunking strategy across both narrative and table queries, making them the safest choices when you cannot tightly control how documents are split.

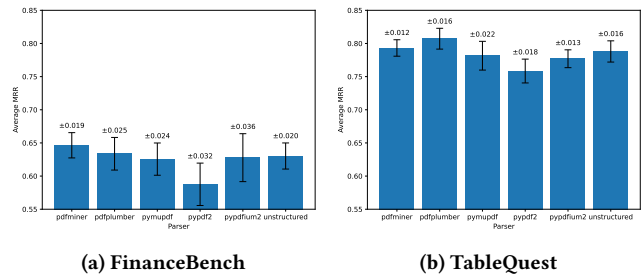

(a) FinanceBench      (b) TableQuest

**Figure 4: Parser stability in page-level MRR (Mean ±1 StdDev, y-axis 0.55–0.85).**

**RQ2c. What is the runtime performance of the different PDF parsers?** We conclude this RQ by also measuring the runtime performance of the six parsers. Table 8 reports wall clock time and the average parsing time per page for TableQuest pages (169 pages). pypdfium2 and PyMuPDF parse a page in well under one second, making them attractive for low-latency pipelines. PyPDF2, pdfminer.six, and pdfplumber require progressively higher costs, from several seconds to tens of seconds, reflecting the overhead of richer layout analysis. Unstructured, which applies full OCR and model-based layout detection using YOLOX [14], is roughly two orders of magnitude slower than the fastest stream parsers.

**RQ2 Short Answer.** pdfminer leads on narrative financial text and pdfplumber on tables, with Unstructured offering a robust middle ground. Pdfminer and Unstructured also exhibit the smallest variance across chunking strategies; however, Unstructured is severely limited by its poor runtime performance. Overall, pdfplumber ranks first across datasets.

Table 8: Parsing times on TableQuest (169 pages).

| Parser | Total Time (s) | Total Time (m:s) | Avg Time/Page (s) |
|---|---|---|---|
| pypdfium2 | 0.46 | 0:00.46 | 0.003 |
| PyMuPDF | 0.62 | 0:00.62 | 0.004 |
| PyPDF2 | 6.04 | 0:06.04 | 0.036 |
| pdfminer | 13.48 | 0:13.48 | 0.080 |
| pdfplumber | 17.54 | 0:17.54 | 0.104 |
| unstructured | 317.23 | 5:17.23 | 1.888 |

### RQ3: How do chunking strategy and overlap percentage affect page-level retrieval performance on financial documents?

**Test:** *We study how chunking strategy and overlap affect page-level retrieval in RAG. First, we fix the retriever to SPLADE, the PDF parser to pdfplumber (the top performer from RQ2, with leading table MRR, near-top narrative-text MRR), and a chunk size of 512 tokens, then compare six chunkers (token, sentence, semantic, recursive, SDPM, neural) using a 25% overlap (128 tokens). Next, in RQ3b, we hold the chunker constant and vary overlap (0%, 25%, 50%) to isolate its influence. In both stages, we evaluate retrieval via P@1, R@3, and MRR.*

**RQ3a. How do chunking strategies affect page-level retrieval?** To address this RQ, we begin by examining overall chunk statistics. For this, we apply *each* chunker to the output of *each* parser. Table 9 summarises the resulting statistics for our default 25% overlap (128 tokens). Totals are aggregated over all six parsers, then separated by dataset, and rows are ordered from the leanest splitter to the most verbose. The recursive chunker yields the fewest segments (about two chunks per page on average) while neural splitting explodes the corpus to more than eight chunks per page, increasing indexing cost by a factor of four. Token and sentence windows sit in the middle, and SDPM and semantic methods roughly double the storage overhead relative to naïve windows.

Table 9: Parser-wise chunk statistics at 25% overlap over 168 FinanceBench and 169 TableQuest pages.

| | FinanceBench | | | | TableQuest | | | |
|---|---|---|---|---|---|---|---|---|
| Chunker | Tot. | Avg | Min | Max | Tot. | Avg | Min | Max |
| Recursive | 2 077 | 2.06 | 1 | 5 | 2 118 | 2.09 | 1 | 7 |
| Token | 2 186 | 2.17 | 1 | 6 | 2 257 | 2.23 | 1 | 9 |
| Sentence | 2 253 | 2.24 | 1 | 10 | 2 270 | 2.24 | 1 | 9 |
| SDPM | 4 954 | 4.91 | 1 | 40 | 5 357 | 5.28 | 1 | 56 |
| Semantic | 5 841 | 5.79 | 1 | 48 | 6 419 | 6.33 | 1 | 59 |
| Neural | 8 129 | 8.06 | 1 | 17 | 8 530 | 8.41 | 1 | 22 |

We now compare the performance of all chunking strategies, while keeping the retriever fixed at SPLADE and the PDF parser at pdfplumber. Table 10 reports the results. Neural leads overall: it achieves the top MRR on both benchmarks (0.658 on FinanceBench, 0.833 on TableQuest). Its advantage over the next alternative is modest on FinanceBench (+0.7 points in MRR vs. sentence) but clearer on TableQuest (+1.4 points vs. sentence and +2.7 vs. token). Neural also leads TableQuest R@3 (0.905), showing that finer-grained splits improve both precision and coverage. Sentence is a solid runner-up. It has the best R@3 on FinanceBench (0.717) and the second-best MRR on both datasets (0.651 / 0.819). On FinanceBench, it ties token for P@1 (0.493) and slightly beats it on MRR (+0.4). This shows that simple context-aware splits remain competitive with low indexing cost. Structure-aware (semantic, SDPM) and recursive trail by 1–5 MRR points. SDPM's strong TableQuest R@3 (0.897) doesn't boost early-rank accuracy. Neural gives the best early-rank results, sentence is a near-optimal, low-cost fallback, and layout-driven splits add size without consistent top-rank gains.

Table 10: Chunking strategies (pdfplumber + SPLADE, 512 tokens, 25% overlap). Best bolded, second-best underlined.

| | FinanceBench | | | TableQuest | | |
|---|---|---|---|---|---|---|
| Chunker | P@1 | R@3 | MRR | P@1 | R@3 | MRR |
| token | <u>0.493</u> | 0.704 | 0.647 | 0.716 | 0.862 | 0.806 |
| sentence | <u>0.493</u> | <u>0.717</u> | <u>0.651</u> | <u>0.733</u> | 0.879 | <u>0.819</u> |
| semantic | 0.447 | 0.671 | 0.606 | 0.690 | 0.871 | 0.795 |
| recursive | 0.480 | <u>0.711</u> | 0.640 | 0.698 | 0.862 | 0.797 |
| SDPM | 0.440 | 0.656 | 0.600 | 0.681 | <u>0.897</u> | 0.794 |
| neural | **0.507** | 0.692 | **0.658** | **0.741** | **0.905** | **0.833** |

**RQ3b. How does chunk overlap (0%, 25%, 50%) impact retrieval using the best chunker?** With neural chunking fixed, a moderate 25% overlap produces the strongest retrieval on both datasets as shown in Table 11. On FinanceBench it lifts MRR from 0.529 (no overlap) to 0.658, and on TableQuest from 0.735 to 0.833, while also giving the highest P@1 and R@3. Doubling the overlap to 50% decreases slightly the performance score (0.647 and 0.832 MRR, respectively), despite doubling the index size, whereas eliminating overlap gives the worst scores. Overall, 25% overlap provides enough context to catch split-boundary queries without the extra storage cost of higher redundancy (refer to Table 9).

Table 11: Effect of chunk overlap on retrieval.

| | FinanceBench | | | TableQuest | | |
|---|---|---|---|---|---|---|
| Overlap | P@1 | R@3 | MRR | P@1 | R@3 | MRR |
| 0 % (0 tokens) | 0.380 | 0.573 | 0.529 | 0.629 | 0.793 | 0.735 |
| 25 % (128 tokens) | **0.507** | 0.692 | **0.658** | **0.741** | **0.905** | **0.833** |
| 50 % (256 tokens) | <u>0.487</u> | <u>0.712</u> | <u>0.647</u> | **0.741** | <u>0.897</u> | <u>0.832</u> |

**RQ3 Short Answer.** The *neural* chunker with a modest 25% overlap yields the best page-level retrieval on both datasets, with *sentence* as a close, lower-cost fallback. Increasing overlap to 50% or using structure-driven chunkers adds index size without consistent accuracy gains.

### RQ4: Do specific combinations of PDF parsers and chunking strategies synergistically enhance performance?

**Test:** *We investigate the interplay between PDF parsers and chunking strategies to determine whether certain combinations lead to improved performance. We conduct a full-factorial experiment, evaluating six parsers (pdfminer, PyMuPDF, PyPDF2, Unstructured, pdfplumber, pypdfium2) with six chunking strategies (token, sentence, semantic, recursive, SDPM, neural). We fix the chunk size at 512 tokens and eliminate overlap (0%) to isolate the interaction between parser and chunking strategy. By analyzing the interaction effects between parsers and chunking methods, we aim to identify optimal* configurations for financial document processing in RAG-based QA systems.

On FinanceBench (Figure 5a), narrative-text retrieval peaks when pairing the recursive chunker with stream parsers: pdfminer + recursive achieves the highest MRR of 0.655, closely followed by pdfplumber + recursive at 0.647. Sentence and token chunkers on pdfminer yield nearly identical MRRs (0.654), underscoring that simple, uniform splits preserve enough context without sophisticated boundary detection. In contrast, layout-agnostic splits like semantic (best 0.568 with pdfminer) and SDPM (best 0.570) fall roughly 8–12 points behind the top performer, showing that tiny, highly fragmented (e.g., 5.79 chunks/page for *semantic* in Table 9) chunks lose the broader context needed for accurate retrieval.

For TableQuest (0% overlap), pdfplumber + sentence leads (MRR 0.804), with pdfplumber + recursive close behind (0.799) and pdfminer + recursive also strong (0.785). Simple splits on pdfplumber stay competitive (e.g., token 0.779), and on pdfminer they remain close (token 0.771, sentence 0.770). Structure-aware chunkers only approach this when paired with layout-sensitive parsers with native table-extraction capabilities: the best they manage: unstructured+semantic (0.776) and unstructured+SDPM (0.772), still trail the leaders by 2–3 points.

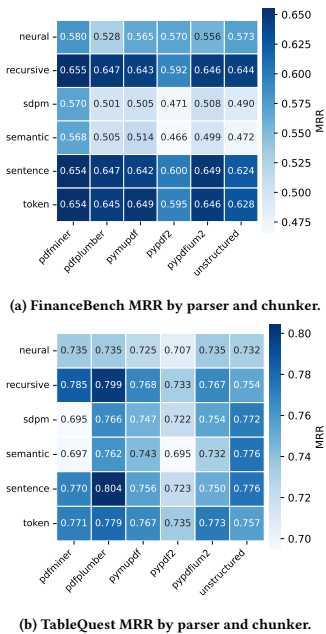

(a) FinanceBench MRR by parser and chunker.

(b) TableQuest MRR by parser and chunker.

Figure 5: Heatmaps of page-level retrieval performance (MRR) by PDF parser and chunking strategy (0% overlap).

**RQ4 Short Answer.** Parser–chunker synergy matters. As simple splits paired with pdfplumber lead, while structure-aware chunkers only approach them when combined with layout-preserving parsers, and still tend to lag.

RQ5: How does LLM size impact the quality and correctness of generated answers in this RAG setting?

**Test:** *We examine how LLM size and type affect response correctness in our RAG pipeline by fixing all other components (SPLADE from RQ1, pdfminer parsing from RQ2, and sentence chunking with 25% overlap from RQ3), while varying only the LLM and the number of retrieved pages (top-1 vs. top-3). We compare a diverse set of small (1.5–4B) and medium (12–27B) open-source models, all hosted locally via Ollama [2], selected for being state-of-the-art and recently released. We also benchmark them against proprietary baselines (GPT-4.1-mini, GPT-5-mini, GPT-5). Answer correctness is automatically assessed using GPT-4o-mini as an LLM-as-judge. This setup enables us to directly measure how model scale and family influence QA accuracy.*

Table 12: Open-source vs. proprietary LLM accuracy (pdfplumber + SPLADE, neural chunks, 25% overlap). $k$ = retrieved pages.

| Model | Size | FinanceBench | | TableQuest | |
|---|---|---|---|---|---|
| | | k = 1 | k = 3 | k = 1 | k = 3 |
| *Small models* | | | | | |
| DeepSeek-R1 | 1.5B | 5.33 | 3.33 | 24.14 | 18.97 |
| Llama-3.2 | 3B | 13.33 | 14.00 | 31.90 | 18.97 |
| Gemma-3 | 4B | 19.33 | 27.33 | 41.38 | 42.24 |
| SmolLM2 | 1.7B | 5.33 | 6.67 | 6.03 | 6.03 |
| Qwen2.5 | 3B | 13.33 | 13.33 | 30.17 | 13.79 |
| *Medium models* | | | | | |
| Gemma-3 | 12B | 28.67 | 36.67 | 42.24 | 43.97 |
| Gemma-3 | 27B | 32.00 | 38.67 | 54.31 | 56.90 |
| Qwen2.5 | 14B | 26.00 | 40.00 | 53.45 | 53.45 |
| Phi-4 | 14B | 32.00 | 38.67 | 47.41 | 55.17 |
| *Proprietary models* | | | | | |
| GPT-4.1-mini | n/a | 34.67 | 52.67 | 57.76 | 67.24 |
| GPT-5-mini | n/a | **36.00** | **54.00** | 59.48 | 68.97 |
| GPT-5 | n/a | 33.33 | 52.00 | 58.62 | 73.28 |

The results in Table 12 highlight important differences between the two benchmarks. On **FinanceBench**, which consists of text-centric financial QA, small open-source LLMs perform poorly, rarely exceeding 20% accuracy, while medium models like Gemma-3 27B and Phi-4 14B reach above 30–38%. Proprietary models provide a further boost, with GPT-5-mini achieving the best FinanceBench result at 54% when retrieving three pages. In contrast, **TableQuest**, which focuses on table-centric QA with varied difficulty levels, is overall more challenging but also more rewarding when stronger models are used: medium open-source LLMs already surpass 50% accuracy, and proprietary models excel, with GPT-5 reaching 73.28% at $k = 3$, the strongest overall score. The gains from increasing the retrieval depth are more pronounced in TableQuest, suggesting that table queries benefit significantly from broader evidence, while narrative queries in FinanceBench show more modest improvements. Overall, the findings confirm that FinanceBench primarily stresses semantic understanding of financial narratives, while TableQuest pushes models to exploit structural cues in tabular data, with proprietary models maintaining a clear edge, but top open-source LLMs narrowing the gap.

**RQ5 Short Answer.** Larger LLMs yield substantially more accurate answers: medium models outperform small ones by roughly 8–15 points in top-1 accuracy.

---

[2]https://ollama.com/

## 6 Related Work

**Financial NLP** is a fast-growing field, that applies NLP to financial data to solve complex challenges and has recently drawn attention for its diverse, impactful applications [6]. Key areas of application within Financial NLP include risk management [16], encompassing tasks like augmenting anti-money laundering investigations, detecting identity fraud via dialogue interactions, and employing deep learning models for bankruptcy prediction using textual disclosures or language modeling for anomaly detection. Another crucial application is management [10], which can involve analysis such as complaint classification for economic and food safety and utilizing pre-trained word embeddings to learn classification lexicons with limited supervision. Furthermore, Financial NLP is extensively used for market sentiment analysis [7], drawing insights from news sentiment and microblog sentiment classification, or analyzing emotional loading in central bank statements and causality analysis of Twitter sentiments and stock market returns. Additionally, the domain focuses on tasks such as financial question answering (QA) [28], where models answer inquiries about product information or assist in financial opinion mining. More recently, the field has seen advancements in the development of pre-trained language models specifically tailored for finance text mining, enhancing the ability to process and understand financial communications [3, 44].

**Financial Question Answering Datasets.** Recent advancements in this field also encompass the construction datasets tailored for finance NLP tasks. Notably, the FinQA [6] and TAT-QA [48] datasets are particularly relevant, as they are designed to require numerical reasoning skills over financial reports that incorporate tabular data for question answering. For instance, FinQA specifically focuses on questions that necessitate locating relevant cells within tabular content and performing numerical operations, such as division, to derive an answer. However, existing hybrid QA datasets, including FinQA and TAT-QA, generally contain only a single flat table per document, thereby lacking examples that demand multi-step reasoning processes across multiple paragraphs and hierarchical tables. To overcome these limitations, the MULTIHIERTT [47] dataset was introduced, distinguishing itself as the first dataset for addressing complicated QA tasks over documents that integrate multiple hierarchical tables and paragraphs. Documents within MULTIHIERTT are characterized by multiple hierarchical tables and longer unstructured texts, requiring a significantly more complex reasoning process across these diverse data types compared to prior benchmarks. This dataset comprises 10,440 question-answer pairs, complete with detailed annotations of reasoning processes and supporting facts.

**RAG Eval.** Evaluation of RAG systems has emerged as a vital research area due to RAG's hybrid retrieval-and-generation architecture. Early work such as RAGAS proposed reference-free metrics for retrieval relevance, generation faithfulness, and system efficiency without relying on human-annotated ground truths [9]. Subsequent benchmarks like RAGBench introduced explainable labels across multiple industry domains under the TRACe framework, enabling large-scale, actionable evaluations on 100 K examples [11]. More recent surveys have synthesized these developments, categorizing metrics by component: precision@k and MRR for retrieval, as well as faithfulness, accuracy, and BLEU-style measures for generation, and highlighting challenges posed by dynamic knowledge sources and lack of unified pipelines [12, 46]. This evolving landscape underscores the ongoing need for integrated, end-to-end evaluation frameworks that seamlessly combine component-level assessments with holistic performance metrics.

## 7 Threats to Validity

**Internal validity.** Our findings may be influenced by specific pipeline settings: fixed 512-token windows, overlap rates of 0%, 25%, and 50%, and a single indexing/inference seed imposed by GPU memory limits (32 GB NVIDIA RTX 5000 Ada). While different window sizes or multiple seeds might change absolute scores, the relative trends are expected to remain stable.

**Construct validity.** Retrieval is judged with P@1, R@3, and MRR, which are standard yet do not capture partial relevance. Answer correctness is scored by a proprietary LLM-as-judge (gpt-4o-mini) rather than human graders; while recent work [27] shows high agreement with expert judgement, automated grading can under- or over-penalise nuanced financial answers. The chosen evaluator also shares architectural biases with the generation models it assesses.

**External validity.** Our study is limited to two finance-oriented datasets (FinanceBench and TableQuest), six open-source parsers, six chunking strategies, four retriever families, and twelve LLMs, including three proprietary models. Results may not generalize to other document domains (e.g., legal or healthcare documents), alternative parsing or chunking tools, or frontier LLMs with substantially different architectures and context lengths. Nonetheless, the chosen components reflect widely used open-source toolkits and representative proprietary baselines, providing a realistic view of practical RAG pipelines for financial QA.

## 8 Conclusion

RAG systems are increasingly adopted to automate the processing of large collections of PDF files in industry. In this paper, we empirically assessed the performance of different parsing and chunking techniques, two critical components of a RAG pipeline. Our evaluation also introduced TableQuest, a benchmark specifically designed to test the handling of tabular data in financial PDFs.

Using FinanceBench and the newly introduced TableQuest benchmark, we derived three key findings: (1) retriever and parser choices strongly affect performance, with lightweight combinations often matching more complex ones, (2) moderate chunk overlaps improve retrieval, while excessive granularity adds cost without consistent benefit, and (3) larger LLMs substantially boost answer correctness, though improvements level off beyond medium scale.

From an industrial perspective, the results from our five RQs provide actionable guidelines for building reliable and efficient RAG pipelines in high-stakes domains such as banking. They demonstrate that lightweight parser–chunker choices already yield strong performance, that moderate chunk overlaps improve retrieval without inflating index size, and that selectively scaling LLMs further improves correctness without prohibitive cost. For practitioners, this translates into better trade-offs between accuracy, runtime, and infrastructure, enabling scalable automation of financial analysis and decision-support workflows.

## Acknowledgments
This research was conducted in the context of an industrial partnership with BGL BNP Paribas and was supported by its industrial funding. This research was also funded in whole or in part by the Luxembourg National Research Fund (FNR), grant reference NCER22/IS/16570468/NCER-FT. We thank the BGL DataLab team for their collaboration and technical input throughout this work. We also thank the anonymous reviewers for their careful reading and constructive feedback.

## References

[1] Narayan S Adhikari and Shradha Agarwal. 2024. A Comparative Study of PDF Parsing Tools Across Diverse Document Categories. *arXiv preprint arXiv:2410.09871* (2024).

[2] Adobe. 2025. What Is a PDF? Portable Document Format. https://www.adobe.com/acrobat/about-adobe-pdf.html. https://www.adobe.com/acrobat/about-adobe-pdf.html Accessed: 14 May 2025.

[3] Dogu Araci. 2019. Finbert: Financial sentiment analysis with pre-trained language models. *arXiv preprint arXiv:1908.10063* (2019).

[4] Artifex Software, Inc. and contributors. 2016. PyMuPDF: High-performance Python PDF library. https://github.com/pymupdf/PyMuPDF. https://github.com/pymupdf/PyMuPDF

[5] Hannah Bast and Claudius Korzen. 2017. A benchmark and evaluation for text extraction from PDF. In *2017 ACM/IEEE joint conference on digital libraries (JCDL)*. IEEE, 1–10.

[6] Zhiyu Chen, Wenhu Chen, Charese Smiley, Sameena Shah, Iana Borova, Dylan Langdon, Reema Moussa, Matt Beane, Ting-Hao Huang, Bryan Routledge, et al. 2021. Finqa: A dataset of numerical reasoning over financial data. *arXiv preprint arXiv:2109.00122* (2021).

[7] Tobias Daudert, Paul Buitelaar, and Sapna Negi. 2018. Leveraging news sentiment to improve microblog sentiment classification in the financial domain. In *Proceedings of the first workshop on economics and natural language processing*. 49–54.

[8] Emily Dinan, Stephen Roller, Kurt Shuster, Angela Fan, Michael Auli, and Jason Weston. 2018. Wizard of wikipedia: Knowledge-powered conversational agents. *arXiv preprint arXiv:1811.01241* (2018).

[9] Shahul Es, Jithin James, Luis Espinosa Anke, and Steven Schockaert. 2024. Ragas: Automated evaluation of retrieval augmented generation. In *Proceedings of the 18th Conference of the European Chapter of the Association for Computational Linguistics: System Demonstrations*. 150–158.

[10] Joao Filgueiras, Luís Barbosa, Gil Rocha, Henrique Lopes Cardoso, Luís Paulo Reis, Joao Pedro Machado, and Ana Maria Oliveira. 2019. Complaint analysis and classification for economic and food safety. In *Proceedings of the Second Workshop on Economics and Natural Language Processing*. 51–60.

[11] Robert Friel, Masha Belyi, and Atindriyo Sanyal. 2024. Ragbench: Explainable benchmark for retrieval-augmented generation systems. *arXiv preprint arXiv:2407.11005* (2024).

[12] Aoran Gan, Hao Yu, Kai Zhang, Qi Liu, Wenyu Yan, Zhenya Huang, Shiwei Tong, and Guoping Hu. 2025. Retrieval Augmented Generation Evaluation in the Era of Large Language Models: A Comprehensive Survey. *arXiv preprint arXiv:2504.14891* (2025).

[13] Yunfan Gao, Yun Xiong, Xinyu Gao, Kangxiang Jia, Jinliu Pan, Yuxi Bi, Yi Dai, Jiawei Sun, Haofen Wang, and Haofen Wang. 2023. Retrieval-augmented generation for large language models: A survey. *arXiv preprint arXiv:2312.10997* 2 (2023).

[14] Zheng Ge, Songtao Liu, Feng Wang, Zeming Li, and Jian Sun. 2021. Yolox: Exceeding yolo series in 2021. *arXiv preprint arXiv:2107.08430* (2021).

[15] Daniel Gozman and Wendy Currie. 2014. The role of investment management systems in regulatory compliance: A post-financial crisis study of displacement mechanisms. *Journal of Information Technology* 29, 1 (2014), 44–58.

[16] Jingguang Han, Utsab Barman, Jer Hayes, Jinhua Du, Edward Burgin, and Dadong Wan. 2018. Nextgen aml: Distributed deep learning based language technologies to augment anti money laundering investigation. Association for Computational Linguistics.

[17] Paul Hopkin. 2018. *Fundamentals of risk management: understanding, evaluating and implementing effective risk management.* Kogan Page Publishers.

[18] Aaron Hurst, Adam Lerer, Adam P Goucher, Adam Perelman, Aditya Ramesh, Aidan Clark, AJ Ostrow, Akila Welihinda, Alan Hayes, Alec Radford, et al. 2024. Gpt-4o system card. *arXiv preprint arXiv:2410.21276* (2024).

[19] Pranab Islam, Anand Kannappan, Douwe Kiela, Rebecca Qian, Nino Scherrer, and Bertie Vidgen. 2023. Financebench: A new benchmark for financial question answering. *arXiv preprint arXiv:2311.11944* (2023).

[20] Gautier Izacard. 2021. PyLate: Python Library for ColBERT-Style Indexing. https://github.com/lightonai/pylate.

[21] Jeremy Singer-Vine and contributors. 2013. pdfplumber: Python PDF Parsing and Extraction Library. https://github.com/jsvine/pdfplumber. https://github.com/jsvine/pdfplumber

[22] Thorsten Joachims and Nick Craswell. 2021. ir-measures: A Library for Evaluating Information Retrieval. https://github.com/terrierteam/ir_measures.

[23] Feyza Duman Keles, Pruthuvi Mahesakya Wijewardena, and Chinmay Hegde. 2023. On the computational complexity of self-attention. In *International Conference on Algorithmic Learning Theory*. PMLR, 597–619.

[24] Omar Khattab and Matei Zaharia. 2020. Colbert: Efficient and effective passage search via contextualized late interaction over bert. In *Proceedings of the 43rd International ACM SIGIR conference on research and development in Information Retrieval*. 39–48.

[25] Carlos Lassance, Hervé Déjean, Thibault Formal, and Stéphane Clinchant. 2024. SPLADE-v3: New baselines for SPLADE. *arXiv preprint arXiv:2403.06789* (2024).

[26] Patrick Lewis, Ethan Perez, Aleksandra Piktus, Fabio Petroni, Vladimir Karpukhin, Naman Goyal, Heinrich Küttler, Mike Lewis, Wen-tau Yih, Tim Rocktäschel, et al. 2020. Retrieval-augmented generation for knowledge-intensive nlp tasks. *Advances in neural information processing systems* 33 (2020), 9459–9474.

[27] Haitao Li, Qian Dong, Junjie Chen, Huixue Su, Yujia Zhou, Qingyao Ai, Ziyi Ye, and Yiqun Liu. 2024. Llms-as-judges: a comprehensive survey on llm-based evaluation methods. *arXiv preprint arXiv:2412.05579* (2024).

[28] Macedo Maia, Siegfried Handschuh, André Freitas, Brian Davis, Ross McDermott, Manel Zarrouk, and Alexandra Balahur. 2018. Www'18 open challenge: financial opinion mining and question answering. In *Companion proceedings of the the web conference 2018*. 1941–1942.

[29] Phillip Massa and Julian McAuley. 2020. rank_bm25: A Python Implementation of Okapi BM25. https://github.com/dorianbrown/rank_bm25.

[30] Norman Meuschke, Apurva Jagdale, Timo Spinde, Jelena Mitrović, and Bela Gipp. 2023. A benchmark of pdf information extraction tools using a multi-task and multi-domain evaluation framework for academic documents. In *International Conference on Information*. Springer, 383–405.

[31] Inc. Ollama. 2024. Ollama: Local LLM Inference Toolkit. https://ollama.com.

[32] pdfminer.six contributors. 2018. pdfminer.six: Python PDF Parsing Library. https://github.com/pdfminer/pdfminer.six. https://github.com/pdfminer/pdfminer.six

[33] Phaseit, Inc. and contributors. 2012. PyPDF2: Pure-Python PDF toolkit. https://github.com/py-pdf/PyPDF2. https://github.com/py-pdf/PyPDF2

[34] pypdfium2-team. 2023. pypdfium2: Python bindings for PDFium. https://github.com/pypdfium2-team/pypdfium2. https://github.com/pypdfium2-team/pypdfium2

[35] Stephen Robertson, Hugo Zaragoza, et al. 2009. The probabilistic relevance framework: BM25 and beyond. *Foundations and Trends® in Information Retrieval* 3, 4 (2009), 333–389.

[36] Daniel Ruffinelli and Joshua Martel. 2022. sparsembed: SPLADE Dense+Sparse Embedding Library. https://github.com/naver/splade.

[37] Zejiang Shen, Ruochen Zhang, Melissa Dell, Benjamin Charles Germain Lee, Jacob Carlson, and Weining Li. 2021. LayoutParser: A Unified Toolkit for Deep Learning Based Document Image Analysis. *arXiv preprint arXiv:2103.15348* (2021).

[38] Gemini Team, Petko Georgiev, Ving Ian Lei, Ryan Burnell, Libin Bai, Anmol Gulati, Garrett Tanzer, Damien Vincent, Zhufeng Pan, Shibo Wang, et al. 2024. Gemini 1.5: Unlocking multimodal understanding across millions of tokens of context. *arXiv preprint arXiv:2403.05530* (2024).

[39] Hugging Face Team. 2023. Transformers: State-of-the-Art Natural Language Processing. https://github.com/huggingface/transformers.

[40] Unstructured Technologies and contributors. 2022. Unstructured: Python library for document preprocessing. https://github.com/Unstructured-IO/unstructured. https://github.com/Unstructured-IO/unstructured

[41] Liang Wang, Nan Yang, Xiaolong Huang, Linjun Yang, Rangan Majumder, and Furu Wei. 2024. Multilingual e5 text embeddings: A technical report. *arXiv preprint arXiv:2402.05672* (2024).

[42] Jason Wei, Xuezhi Wang, Dale Schuurmans, Maarten Bosma, Fei Xia, Ed Chi, Quoc V Le, Denny Zhou, et al. 2022. Chain-of-thought prompting elicits reasoning in large language models. *Advances in neural information processing systems* 35 (2022), 24824–24837.

[43] Peng Xu, Wei Ping, Xianchao Wu, Lawrence McAfee, Chen Zhu, Zihan Liu, Sandeep Subramanian, Evelina Bakhturina, Mohammad Shoeybi, and Bryan Catanzaro. 2023. Retrieval meets long context large language models. In *The Twelfth International Conference on Learning Representations*.

[44] Yi Yang, Mark Christopher Siy Uy, and Allen Huang. 2020. Finbert: A pretrained language model for financial communications. *arXiv preprint arXiv:2006.08097* (2020).

[45] Antonio Jimeno Yepes, Yao You, Jan Milczek, Sebastian Laverde, and Renyu Li. 2024. Financial report chunking for effective retrieval augmented generation. *arXiv preprint arXiv:2402.05131* (2024).

[46] Hao Yu, Aoran Gan, Kai Zhang, Shiwei Tong, Qi Liu, and Zhaofeng Liu. 2024. Evaluation of retrieval-augmented generation: A survey. In *CCF Conference on Big Data*. Springer, 102–120.

[47] Yilun Zhao, Yunxiang Li, Chenying Li, and Rui Zhang. 2022. MultiHiertt: Numerical reasoning over multi hierarchical tabular and textual data. *arXiv preprint arXiv:2206.01347* (2022).

[48] Fengbin Zhu, Wenqiang Lei, Youcheng Huang, Chao Wang, Shuo Zhang, Jiancheng Lv, Fuli Feng, and Tat-Seng Chua. 2021. TAT-QA: A question answering benchmark on a hybrid of tabular and textual content in finance. *arXiv preprint arXiv:2105.07624* (2021).